\documentclass[lettersize,journal]{IEEEtran}

\usepackage{graphicx} 
\usepackage{xcolor}
\usepackage{booktabs}
\usepackage{colortbl}
\usepackage{makecell}
 \usepackage{url}

\title{VAD4Space: Visual Anomaly Detection\\ for Planetary Surface Imagery}

\author{
Fabrizio Genilotti\textsuperscript{1},
Arianna Stropeni\textsuperscript{1},
Francesco Borsatti\textsuperscript{1},
\\
Manuel Barusco\textsuperscript{1},
Davide Dalle Pezze\textsuperscript{1},
Gian Antonio Susto\textsuperscript{1} \\[6pt]
\textsuperscript{1}University of Padova, Italy \\
\texttt{fabrizio.genilotti@studenti.unipd.it}, 
\texttt{arianna.stropeni@studenti.unipd.it}, 
\texttt{francesco.borsatti@phd.unipd.it},
\texttt{manuel.barusco@phd.unipd.it},
\texttt{davide.dallepezze@unipd.it},
\texttt{gianantonio.susto@unipd.it}
}

\begin{document}
\maketitle

\begin{abstract}
Space missions generate massive volumes of high-resolution orbital and surface imagery that far exceed the capacity for manual inspection. 
Of particular scientific interest is the automated detection of rare phenomena, which may represent valuable discoveries or previously unanticipated surface processes. 
Traditional supervised learning approaches are fundamentally limited in this context by the scarcity of labeled rare events and by closed-world assumptions that preclude the discovery of truly novel observations.
In this work, we investigate Visual Anomaly Detection (VAD) as a framework for automated discovery in planetary exploration. We present the first empirical evaluation of state-of-the-art feature-based VAD methods on real planetary imagery, encompassing both orbital lunar data and Mars rover surface imagery. 
To support this evaluation, we introduce two benchmarks: (i) a lunar dataset derived from Lunar Reconnaissance Orbiter Camera Narrow Angle imagery, comprising of fresh and degraded craters as anomalies alongside normal terrain; and (ii) a Mars surface dataset designed to reflect the characteristics of rover-acquired imagery.
We evaluate multiple VAD approaches with a focus on computationally efficient, edge-oriented solutions suitable for onboard deployment, applicable to both orbital platforms surveying the lunar surface and surface rovers operating on Mars. 
Our results demonstrate that feature-based VAD methods can effectively identify rare planetary surface phenomena while remaining feasible for resource-constrained environments. By grounding anomaly detection in planetary science, this work establishes practical benchmarks and highlights the potential of open-world perception systems to support a range of mission-critical applications, including tactical planning, landing site selection, hazard detection, bandwidth-aware data prioritization, and the discovery of unanticipated geological processes.
\end{abstract}

\section{Introduction}
\label{sec:intro}

Space exploration represents one of the most advanced scientific and technological frontiers of our time. 
Modern missions generate data at a scale that fundamentally outpaces human capacity to review it, making automated analysis not merely a convenience but an operational necessity.
From lunar orbit, the Lunar Reconnaissance Orbiter Camera (LROC) has accumulated high-resolution imagery covering the lunar surface. On Mars, rovers such as Curiosity and Perseverance produce continuous streams of multispectral imagery and telemetry, from which scientists must extract actionable insight under severe operational constraints: data exchange between Mars and Earth is possible only during brief communication windows aligned with orbital geometry, and every planning cycle must count.
\\
The dominant response to this data volume has been supervised deep learning e.g. image classification, which has proven effective across a range of planetary science tasks, from crater detection to terrain classification. However, this paradigm carries two fundamental limitations that are particularly critical in the space domain.
First, producing reliable annotations demands domain expertise and is prohibitively time-consuming at scale. 
Second, and more critically, the phenomena of greatest scientific interest are almost by definition rare. 
Fresh impact craters, rockfalls, slope failures, mineral veins, and meteorites occupy only a tiny fraction of any image archive, making it practically impossible to assemble the labeled examples that supervised methods require. 
More fundamentally, supervised approaches operate under the closed-world assumption, the implicit requirement that all relevant categories be known and anticipated at training time, which is unfeasible in planetary exploration, where discovering the unexpected is often a primary scientific goal.
\\
Visual Anomaly Detection (VAD) is an emerging field that offers an alternative to supervised approaches with potential in many vision applications.
VAD models are trained exclusively on abundant normal data and, at inference time, identify anything that deviates from the expected normal distribution, without requiring a single anomalous example during training and without presupposing what form anomalies may take.
This open-world property is precisely what planetary exploration often demands. 
Beyond detection, VAD models produce pixel-level anomaly maps that localize suspicious regions within each image.
This provides human operators with direct visual evidence, enabling rapid, informed decision-making without requiring expert re-analysis of raw imagery.
This capacity for human-in-the-loop interaction is particularly valuable in mission contexts where bandwidth is limited and operator time is precious. 
Finally, feature-based VAD methods can be made computationally lightweight, making them viable candidates for onboard deployment on resource-constrained platforms such as orbital satellites and surface rovers.
\\
Despite this potential, VAD has been developed and benchmarked almost exclusively for industrial inspection and medical imaging. Its applicability to planetary science remains largely unexplored, and no systematic evaluation of real planetary imagery exists to date. This paper addresses that gap directly.
\\
Our main contributions are as follows:
\\
\textbf{(i) First empirical evaluation of VAD on planetary imagery}. We systematically benchmark seven feature-based VAD methods on real orbital and surface planetary data, demonstrating that open-world anomaly detection is both feasible and effective in the space exploration domain.
\\
\textbf{(ii) Two new planetary VAD benchmarks}. We introduce a lunar benchmark derived from Lunar Reconnaissance Orbiter Camera Narrow Angle imagery and a Mars surface benchmark built from Curiosity rover Mastcam data, providing the community with standardized evaluation protocols for anomaly detection in planetary science.
\\
\textbf{(iii) A three-setting evaluation protocol}. Beyond the standard balanced and rare-anomaly settings, we introduce a training contamination setting that better reflects real deployment conditions, where a perfectly clean training set cannot be guaranteed. This setting reveals important differences in model robustness that would be invisible under idealized assumptions.
\\
\textbf{(iv) Edge-oriented analysis}. We use lightweight backbones during the experiments, demonstrating that strong detection performance is achievable with reduced memory and compute, making onboard deployment on resource-constrained orbital and surface platforms practically viable.
\\
The remainder of this paper is organized as follows. Section \ref{sec:related_work} reviews related work on VAD and Novelty Detection in the space domain. 
Section \ref{sec:methodology} describes the planetary datasets and VAD methods evaluated. Section \ref{sec:experimental_setting} outlines the experimental settings and evaluation metrics. Section \ref{sec:results} presents and discusses the results. Finally, Section \ref{sec:conclusion} draws conclusions and outlines future directions.

\begin{figure}[t]
\centering
\includegraphics[width=\linewidth]{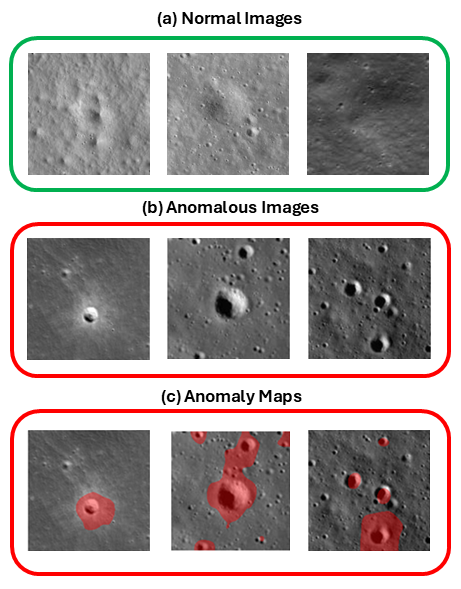}
\caption{ Images from the Lunar Dataset.
(a) Representative normal terrain images. 
(b) Anomalous images.
 (c) Corresponding anomaly maps produced by PaDiM overlaid on the anomalous images, 
where red regions highlight the areas identified as anomalous. 
Despite being trained exclusively on normal terrain, PaDiM 
successfully localizes crater structures without any supervision.}
\label{fig:lunar_anomaly_maps}
\end{figure}

\section{Related Work}
\label{sec:related_work}

\subsection{Visual Anomaly Detection}

VAD plays a crucial role in many computer vision applications, including manufacturing and healthcare \cite{MVTec,bao2024bmad}. VAD models offer two main advantages. First, they are typically trained in an unsupervised manner using only normal samples, bypassing the costly process of collecting and labelling large numbers of defective examples. Second, they produce pixel-level anomaly maps that enhance interpretability, operator decision-making, and end-user trust.
\\
State-of-the-art VAD models generally fall into two main categories: reconstruction-based methods and feature embedding–based methods.
\\
\textbf{Reconstruction-based methods} employ generative models to learn the distribution of normal data, identifying anomalies at inference time through high reconstruction errors. Common approaches include Autoencoders, GANs, and Diffusion Models. However, these methods are often computationally expensive, which can limit their applicability in real-time or resource-constrained scenarios.
\\
\textbf{Feature embedding–based methods} exploit representations extracted from pretrained neural networks, avoiding explicit image reconstruction and achieving higher computational efficiency.
This family is further organized into three subcategories: \textbf{Teacher–Student methods}, which detect anomalies through discrepancies between teacher and student feature maps (e.g. STFPM \cite{st_pyramid}); \textbf{Memory Bank methods}, which store normal feature representations for comparison at inference time (e.g., PaDiM \cite{PaDiM}, PatchCore \cite{patch}, CFA \cite{lee2022cfa}); and \textbf{Normalizing Flow methods}, which map data distributions to a normal distribution for likelihood-based anomaly detection (e.g. FastFlow \cite{yu2021fastflow}).

\subsection{Novelty Detection for Planetary Imagery in Space Exploration}

Planetary missions generate vast scientific data volumes that challenge manual analysis. 
Impact crater detection is one of the most studied problems in planetary image analysis, given the central role craters play in the morphological evolution of the surfaces of planets.
\\
LROC's Narrow Angle Cameras provide 5-km swaths and small image chipouts centered on regions of potential scientific interest are extracted, enabling detection of craters, rockfalls, and volcanic deposits \cite{robinson2010lunar}.
Distinguishing fresh from degraded craters is stratigraphically important: fresh craters retain ejecta while older ones lose it through regolith overturn \cite{dunkel2022lrocnet}.
Building directly on the foundations described above, \cite{dunkel2022lrocnet} introduces LROCNet, a deep learning classifier designed to enable content-based search of lunar imagery in NASA's Planetary Data System.
They achieved ~82\% accuracy on this task using CNNs trained on 5,000 LROC chipouts. 
Specifically, VGG-11 achieves the best performance and is selected for deployment. 
\\
However, the availability of labeled training data remains a fundamental bottleneck in supervised learning for planetary science applications. Expert annotation is costly and time-intensive, particularly when dealing with millions of image patches. 
The system provides classification information for craters; a large number of geologically and geophysically significant structures are entirely absent from the classification framework, including  volcanic constructs, landslides, and rockfalls.
\\
\\
Authors of \cite{kerner2020comparison} address the problem of novelty detection in multispectral images acquired by the Mastcam instrument onboard the Mars Science Laboratory (MSL) Curiosity rover, with the ultimate goal of accelerating tactical science planning during rover-based planetary exploration missions. 
The work is motivated by the severe time constraints faced by MSL science teams, who typically have less than 12 hours to review newly downlinked data and plan follow-up observations. 
In this context, the authors argue that automated novelty detection systems capable of rapidly prioritizing the most scientifically interesting observations could significantly increase the scientific return of such missions, provided that they also offer interpretable explanations that allow scientists to understand and trust the system's outputs.
\\
The authors of \cite{kerner2020comparison} compare several models, focusing on autoencoder approaches.
\textbf{Convolutional Autoencoder (CAE)} \cite{masci2011stacked} is trained exclusively on typical examples and assesses novelty via reconstruction error: the underlying intuition is that a model trained on typical images will fail to accurately reconstruct novel ones, yielding higher errors for out-of-distribution inputs \cite{thompson2002implicit}. 
\\
However, deep generative models generalize so effectively that they can faithfully reconstruct even anomalous regions, meaning the reconstruction error between input and output remains low regardless of whether the input contains a defect. This undermines the core assumption that anomaly regions will produce high reconstruction errors. 
\\
\\
To address these limitations, recent advances in the VAD literature have shifted toward feature embedding-based approaches that completely bypass image reconstruction. Instead of modeling pixels directly, these methods learn discriminative feature representations, enabling robust and efficient anomaly detection, achieving excellent performance across many industrial and medical tasks.
In this work, we take an important step toward closing this gap in the context of planetary exploration. Specifically, we perform a comprehensive evaluation of several state-of-the-art feature-based VAD methods using real planetary imagery. Our objective is to rigorously determine whether the exceptional performance these methods achieve in other domains translates effectively to the unique visual characteristics and operational challenges of space exploration environments.

\section{Methodology}
\label{sec:methodology}

A central goal of this work is to provide the first empirical evaluation of Visual Anomaly Detection on real planetary imagery. To this end, we introduce two new benchmarks grounded in operational space exploration scenarios: one targeting anomalous surface features in lunar orbital imagery, described in Section \ref{subsec:lunar_methodology}, and one focused on geologically novel observations in Mars rover surface imagery, described in Section \ref{subsec:mars_methodology}. 
Both benchmarks are evaluated under three evaluation protocols of increasing realism, described in Section \ref{sec:evaluation_settings}.
Section \ref{subsec:vad_methods} then presents the seven feature-based VAD methods considered in this study.

\begin{figure}[t]
\centering
\includegraphics[width=\linewidth]{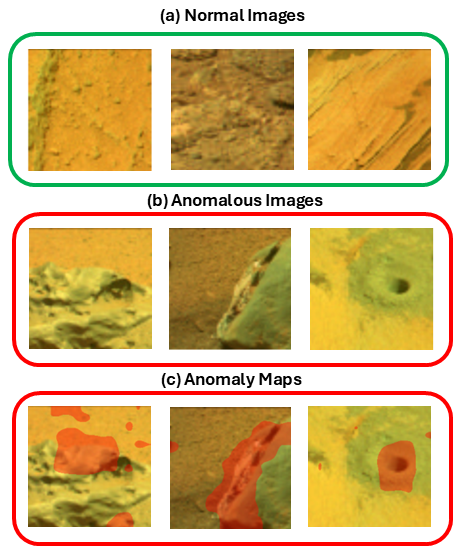}

\caption{ Images from the Mars Dataset.
(a) Representative normal terrain images, 
(b) Anomalous images
(c) Corresponding anomaly maps produced by PaDiM overlaid on the anomalous images, 
where red regions highlight the areas identified as anomalous. 
}
\label{fig:mars_anomaly_maps}
\end{figure}

\subsection{Lunar Benchmark}
\label{subsec:lunar_methodology}

\noindent \textbf{Motivation}
From an operational standpoint, automated anomalous surface detection is increasingly critical for mission planning, as it can help landers and rovers find safe, obstacle-free terrain. 
Identifying recent rockfalls, slope failures, or newly formed craters in candidate landing zones, all detectable in orbital imagery, can directly inform site selection and hazard avoidance strategies. 
\\
\noindent However, spacecraft operating far from Earth cannot transmit everything they observe; bandwidth constraints force onboard prioritization decisions about what data is worth sending home at all. 
This motivates performing onboard anomaly detection, where the spacecraft itself identifies novel or unexpected observations and prioritizes their downlink.

\noindent \textbf{Dataset Description}
The dataset is composed of three mutually exclusive classes: \textit{fresh craters (ejecta), old craters, and none} \cite{emily_dunkel_2022_7041842}.
\textbf{Fresh craters} are characterized by the presence of impact ejecta, material displaced outward by the meteorite impact, which appears as a bright, often radially symmetric halo surrounding the crater bowl. 
\\
\textbf{Old craters} lack visible ejecta and are sometimes referred to as degraded craters. 
\\
\textbf{None class} represent the majority of the images, i.e., they contain no qualifying surface feature.
This strong class imbalance reflects the true distribution of the lunar surface: most of it, at any given location and resolution, presents no particularly salient geological feature. 
In this work, we treat the none class (majority class) as the normal class, and the two minority classes, fresh craters and old craters, as the anomalous classes.
\\
The dataset used consists of image chipouts derived from imagery captured by the Lunar Reconnaissance Orbiter Camera (LROC) Narrow Angle Cameras \cite{dunkel2022lrocnet}. 
It consists of 5,000 labeled image chipouts (small cutouts) extracted from full 5-km swath images. The full dataset is publicly available on Zenodo \footnote{\url{https://zenodo.org/records/7041842}}.
The labeling process began with 7 candidate classes:\textit{ fresh crater (with impact ejecta), old crater, multiple overlapping old craters, irregular mare patches, rockfalls and landslides, of scientific interest, and none}. Due to the rarity of examples in most of the minority classes, the final dataset was reduced to three classes

\subsection{Mars Benchmark}
\label{subsec:mars_methodology}

\noindent \textbf{Motivation}
The MSL Curiosity rover operates on a tight "tactical planning" cycle: the rover collects data, transmits it to Earth, scientists analyze it, and send new commands.
This process must happen all within a narrow window when a clear line of sight exists, dictated by the alignment of Mars, Earth, and the Deep Space Network antennas. The core challenge, then, is one of time and distance.
\\
This is where Visual Anomaly Detection (VAD) becomes critical. 
Because transmission bandwidth is limited and scientists cannot realistically review every image the rover captures, the rover must prioritize and transmit only the most valuable data.
VAD addresses this by performing novelty detection: automatically identifying the observations that deviate most from the expected, and surfacing them for human review.
\\
Most images of the Martian surface look broadly similar, composed of \textit{dust, rocks, and typical sediment}, but genuinely scientifically valuable observations, such as \textit{meteorites, exposed fresh rock interiors, or drill tailings}, are rare and highly informative. A VAD system trained on typical Martian geology can automatically flag these unusual cases and push them to the top of the review queue, allowing scientists to concentrate their limited time on the most promising targets rather than scrolling through hundreds of routine images. The result is a faster, more informed decision-making process built around the constraints of interplanetary exploration.

\noindent \textbf{Dataset Description}
The dataset was built from 477 multispectral (6-band) images acquired by the Mastcam M-100 (right eye) camera onboard the Curiosity rover, covering sols 1 to 1666 (August 6, 2012 to April 14, 2017). The right eye was chosen simply because it produced more multispectral images than the left eye during that period. All images use six narrow-band spectral filters spanning roughly 400–1100 nm (visible to near-infrared).
\\
Two domain experts manually reviewed all 477 images and drew bounding boxes around regions they considered geologically novel. 
They identified 237 novel bounding boxes across 156 images, while the remaining 321 images were classified as containing only typical geology.
To increase the number of available training samples, the authors applied a sliding window approach, extracting 64×64×6 pixel tiles \cite{hannah_kerner_2018_3732485}.
The training set consists of 9,302 typical tiles, while the test set comprises 426 typical tiles and 430 novel tiles (856 total).
The 430 novel test tiles are divided into 8 geological categories: \textit{DRT spot, dump pile, broken rock , drill hole, meteorite, vein, float rock, and bedrock.}
The complete dataset is publicly available on Zenodo  \footnote{\url{https://zenodo.org/records/3732485}}.
\\
Following the standard one-class classification setup for novelty detection, models are trained exclusively on typical geology samples and evaluated against both typical and novel tiles, where the latter serve as the anomalous class.
\\
The novelties in the dataset are not all of the same kind; some are primarily \textbf{spectral novelties} (the material has an unusual reflectance spectrum, like a meteorite or a vein), while others are primarily \textbf{morphological novelties} (the spatial pattern is unusual, like a drill hole or a DRT spot). 
The core issue is that RGB images collapse the full spectral information into just three broad channels. For spectrally novel classes like meteorites and veins, the novelty often manifests as subtle differences in reflectance at specific wavelengths, particularly in the near-infrared range (up to 1100 nm) that is completely invisible to a standard RGB camera. 
\\
It should be noted that while the original experiments in \cite{kerner2020comparison} leveraged all six narrow-band spectral filters from the Mars images, only the three RGB channels are used in our experiments, ignoring the remaining spectral information. 
This is due to feature-based models that use pretrained backbones, which exploit only RGB channels.
The implications of this are discussed further in Section \ref{sec:results}.

\subsection{Evaluation Settings}
\label{sec:evaluation_settings}

We evaluate all VAD models under three distinct experimental settings for LROC and Mars datasets, which are designed to progressively move from idealized benchmarking conditions toward realistic deployment scenarios.
\\
\textbf{Setting 1 – Balanced Test Set:}
It follows the original experimental protocol of each dataset, where the test set contains a roughly balanced distribution of normal and anomalous samples. While this allows direct comparison with the results reported in the original dataset papers, it does not reflect the true operating conditions of anomaly detection systems: in practice, anomalies are rare by definition.
\\
\textbf{Setting 2 – 5\% Positives:} This setting adopts the standard VAD evaluation protocol. The models are trained exclusively on normal samples, and the test set is constructed such that anomalous images constitute only 5\% of the total, with the remaining 95\% being normal. This reflects the core assumption of anomaly detection, that anomalies are rare events, and is the canonical setting studied in the VAD literature.
\\
\textbf{Setting 3 – 5\% Positives with Contamination: } 
It extends Setting 2 to model a fully realistic deployment scenario that is rarely considered in the VAD literature or in the original dataset papers. Here, the training set is no longer assumed to be perfectly clean: a small fraction (5\%) of anomalous samples are present among the training data, reflecting the practical difficulty of curating a purely normal training set in space exploration contexts, where all collected data must be labeled by domain experts. This contamination introduces noise into the learned normal distribution, making the detection task substantially harder.

\subsection{VAD Methods}
\label{subsec:vad_methods}

As discussed in Section \ref{sec:related_work}, VAD methods fall into two main categories: reconstruction-based approaches using image-to-image generative models, and feature-based approaches exploiting representations from pre-trained networks.
Since both scenarios, orbital lunar imagery and Mars rover data, assume edge processing (onboard the satellite or rover), computational efficiency is a primary concern. We therefore focus on feature-based VAD methods, which can be readily adapted to edge deployments by swapping the backbone \cite{barusco2024paste}. 
Specifically, we replace the standard WideResNet50 \cite{zagoruyko2016wide} with a lightweight MobileNetV2 \cite{sandler2018mobilenetv2}, which substantially reduces parameters and operations while retaining sufficient representational power for anomaly detection (see Table \ref{tab:models_profile}).
In the following, we briefly describe the seven feature-based VAD methods considered in this study:
\begin{itemize}
    \item \textbf{Patchcore}: it builds a compact memory bank of representative normal patches and flags anomalies based on the distance of test patches to their nearest neighbors \cite{patch}.
    \item \textbf{Padim}: it models each spatial feature location with a multivariate Gaussian and uses the Mahalanobis distance to detect deviations as anomalies \cite{PaDiM}.
    \item \textbf{CFA}: it creates a memory of normal patch embeddings and adapts them into coupled hyperspheres to amplify the separation between normal and abnormal feature representations \cite{lee2022cfa}.
    \item \textbf{STFPM}: it is based on two networks (teacher and student) with knowledge distillation, where student and teacher feature map deviations indicate anomalies \cite{st_pyramid}.
    \item \textbf{RD4AD}: it improves the STFPM approach by considering an autoencoder-like approach where the teacher is the encoder and the student the decoder \cite{rd4ad}.
    \item \textbf{SimpleNet}: it also uses a feature adaptor like CFA, but also considers the generation of synthetic anomalies at feature-level to improve performance  \cite{rolih2025supersimplenet}.
    \item \textbf{FastFlow}: it is based on normalizing flow models to transform complex input data distributions into normal distributions, leveraging probability as a measure of normality \cite{yu2021fastflow}.    
\end{itemize}

\section{Experimental Setting}
\label{sec:experimental_setting}

\begin{table*}[]
\centering
\caption{Anomaly detection results on the Lunar dataset under three evaluation protocols (Fully, 5\% Positives, and 5\% Positives+Contamination) discussed in Section \ref{sec:evaluation_settings}, showing all metrics reported in Section \ref{subsec:evaluation_metrics}: image-level AUROC (img-ROC), F1 score (img-F1), and area under the Precision-Recall curve (AUC-PR). Bold values indicate the best-performing model for each metric and setting.}
\resizebox{\textwidth}{!}{%
\begin{tabular}{l c >{\columncolor{gray!12}}c c >{\columncolor{gray!12}}c c >{\columncolor{gray!12}}c c c >{\columncolor{gray!12}}c c}
\toprule
\rowcolor{white} & \multicolumn{4}{c}{Full} & \multicolumn{3}{c}{5\% Positives} & \multicolumn{3}{c}{5\% Positive + Contamination} \\
\cmidrule(lr){2-5} \cmidrule(lr){6-8} \cmidrule(lr){9-11}
\rowcolor{white}Model & img-ROC & img-F1 & AUC-PR & img-ACC & img-ROC & img-F1 & AUC-PR & img-ROC & img-F1 & AUC-PR \\
\midrule
Padim     & 0.68 & \textbf{0.53} & \textbf{0.52} & \textbf{0.73} & \textbf{0.71} & \textbf{0.28} & \textbf{0.16} & 0.63 & 0.21 & 0.14 \\
CFA       & 0.66 & 0.52 & 0.45 & 0.71 & 0.69 & 0.21 & 0.11 & 0.63 & 0.18 & 0.08 \\
RD4AD     & 0.52 & 0.49 & 0.32 & 0.69 & 0.58 & 0.14 & 0.07 & 0.47 & \textbf{0.48} & \textbf{0.29} \\
SimpleNet & 0.51 & 0.48 & 0.33 & 0.69 & 0.51 & 0.14 & 0.06 & 0.46 & 0.11 & 0.05 \\
PatchCore & 0.62 & 0.47 & 0.40 & 0.59 & 0.65 & 0.21 & 0.11 & \textbf{0.64} & 0.18 & 0.07 \\
STFPM     & 0.58 & 0.47 & 0.32 & 0.52 & 0.59 & 0.15 & 0.07 & 0.45 & 0.12 & 0.05 \\
FastFlow  & \textbf{0.70} & 0.52 & 0.47 & 0.70 & 0.67 & 0.17 & 0.11 & 0.61 & 0.23 & 0.13 \\
\bottomrule
\end{tabular}}
\label{tab:results_lunar}
\end{table*}

\subsection{Evaluation Metrics}
\label{subsec:evaluation_metrics}

We evaluate both methods using standard metrics for visual anomaly detection:

\begin{itemize}
\item \textbf{Image-level AUROC (I-ROC)}: Area under the Receiver Operating Characteristic Curve for classifying images as normal or anomalous.
\item \textbf{Image-level PR AUROC (AUC-PR)}:   Area under the Precision-Recall Curve summarizes the trade-off between precision and recall.
\item \textbf{Accuracy}: Proportion of correctly classified images over the total number of test images. While accuracy is generally not a meaningful metric in anomaly detection due to the inherent class imbalance, we report it exclusively for Setting 1 to enable a direct comparison with the supervised classification results reported in the original lunar dataset paper.
\item \textbf{Memory Footprint}: Total memory required to store model parameters and auxiliary data structures (memory bank, statistics, etc.).
\item \textbf{Inference Time}: Average time required to process a single test image on a target edge device.
\item \textbf{Inference Peak memory}: Average peak memory needed to process a single test image on a target edge device.
\end{itemize}

While in VAD literature, metrics like AUROC are also calculated at pixel-level, unfortunately, the identified dataset did not provide the regions inside the image where the class/anomaly was present.
Therefore, we provide in Figure \ref{fig:lunar_anomaly_maps} and \ref{fig:mars_anomaly_maps} anomaly maps produced by the VAD models that give an intuition of why an image was flagged as anomalous.

\subsection{Implementation Details}
For all the VAD methods evaluated in this study, the chosen backbone is MobileNetV2 for feature extraction, as it provides a good balance between representational power and computational efficiency \cite{barusco2024paste}.
Features are extracted from three intermediate layers to capture multi-scale semantic information at different spatial resolutions. 
The latency and peak memory results reported in Table~\ref{tab:models_profile} were measured on a CPU-only platform equipped with an Intel Core i5-9600K processor clocked at 3.70 GHz, reflecting the computational constraints typical of edge-deployable hardware.

\section{Results}
\label{sec:results}
We report the experimental results obtained by evaluating the VAD models on both benchmarks. Sections~\ref{ssec:comparison_lunar} and~\ref{ssec:comparison_mars} present the performance comparisons under Setting 1 for the Lunar and Mars benchmarks, respectively.
Section~\ref{subsec:realistic_deployment_setting} discusses Setting 3, the most realistic deployment configuration for both datasets. Finally, Section~\ref{subsec:qualitative_results} examines the qualitative results of the interpretations produced by the VAD models.

\begin{table*}[]
\centering
\caption{Anomaly detection results on the Mars dataset under three evaluation protocols (Fully, 5\% Positives, and 5\% Positives+Contamination) discussed in Section \ref{sec:evaluation_settings}, showing all metrics reported in Section \ref{subsec:evaluation_metrics}, image-level AUROC (img-ROC), F1 score (img-F1), and area under the Precision-Recall curve (AUC-PR). Bold values indicate the best-performing model for each metric and setting.
}
\begin{tabular}{l c >{\columncolor{gray!12}}c c >{\columncolor{gray!12}}c c >{\columncolor{gray!12}}c c >{\columncolor{gray!12}}c c}
\toprule
\rowcolor{white} & \multicolumn{3}{c}{Full} & \multicolumn{3}{c}{5\% Positives} & \multicolumn{3}{c}{5\% Positives + Contamination} \\
\cmidrule(lr){2-4} \cmidrule(lr){5-7} \cmidrule(lr){8-10}
\rowcolor{white}Model & img-ROC & img-F1 & AUC-PR & img-ROC & img-F1 & AUC-PR & img-ROC & img-F1 & AUC-PR \\
\midrule
Padim     & 0.70 & 0.72 & 0.68 & 0.73 & 0.22 & 0.12 & \textbf{0.71} & \textbf{0.22} & 0.10 \\
CFA       & 0.68 & 0.68 & 0.61 & 0.68 & 0.22 & 0.09 & 0.62 & 0.15 & 0.07 \\
RD4AD     & 0.66 & 0.73 & 0.65 & 0.66 & 0.22 & 0.12 & 0.65 & 0.20 & 0.10 \\
SimpleNet & 0.60 & 0.69 & 0.59 & 0.61 & 0.16 & 0.10 & 0.62 & 0.19 & 0.12 \\
PatchCore & \textbf{0.81} & \textbf{0.79} & \textbf{0.75} & \textbf{0.80} & \textbf{0.25} & \textbf{0.16} & 0.40 & 0.11 & 0.04 \\
STFPM     & 0.50 & 0.51 & 0.51 & 0.51 & 0.11 & 0.05 & 0.70 & 0.20 & \textbf{0.13} \\
FastFlow  & 0.72 & 0.71 & 0.70 & 0.66 & 0.19 & 0.12 & 0.67 & 0.21 & \textbf{0.13} \\
\bottomrule
\end{tabular}
\label{tab:results_mars}
\end{table*}

\subsection{Performance Comparison in Lunar Benchmark}
\label{ssec:comparison_lunar}

In Table \ref{tab:results_lunar} we report anomaly detection results on the LROC dataset with the metrics defined in Section \ref{subsec:evaluation_metrics}. 
\\
Under Setting 1, which mirrors the original paper's balanced test setup, our best model, PaDiM, achieves 73\% img-ACC versus the supervised baseline of ~82\%, a gap of less than 10\% with no label supervision during training.
\\
In addition, unlike supervised models, VAD can detect anomaly classes absent from training, such as rockfalls, which were excluded from the original dataset due to insufficient samples.
Since VAD flags any deviation from the learned distribution, it naturally generalizes to these rare but scientifically significant phenomena.
\\
Moreover, PaDiM requires only 10.7 MB of storage versus the 507 MB required by the VGG-11 model used in the original paper (Table~\ref{tab:models_profile}). 
Therefore, VAD models are well-suited for onboard deployment for integration into orbital platforms tasked with autonomous lunar surface surveying.

\subsection{Performance Comparison in Mars Benchmark}
\label{ssec:comparison_mars}

Table \ref{tab:results_mars} presents the results across all evaluation settings. 
Examining Setting 1 (Balanced Test set), the best baseline from the original paper, CAE-MSE, achieves an IMG-ROC of 0.705, while our best model, PatchCore, reaches 0.81, a substantial improvement, and notably using only 3 RGB channels versus the original 6. Although the additional channels carry meaningful information for certain anomaly types, this trade-off is inherent to using pretrained backbones designed for standard RGB input.
\\
In terms of model size, we estimate the original baselines at approximately 34 MB, whereas PaDiM and PatchCore require only 10.7 MB and 17.4 MB, respectively. Our backbones also natively support up to 256×256 resolution, compared to the 64×64 used in the original work.
\\
Examining performance and memory, VAD models show to be more accurate and lighter, making them suited for edge deployment for Mars rover operations.

\subsection{Realistic Deployment Setting}
\label{subsec:realistic_deployment_setting}

Before moving to Setting 3, it is worth noting that metrics shift slightly between Settings 1 and 2 for both benchmarks.
Overall performance remains comparable since the training data are identical, and the only difference is a test set that is calibrated to better reflect the true distribution, where anomalies are rare.
\\
Setting 3 is the most realistic and challenging, as it abandons the classical assumption of VAD literature (and adopted by the original work on the Mars Dataset) that training sets contain only normal samples. Instead, a small fraction of anomalies (5\%) is present during training, reflecting better real-world conditions.
\\
On the Lunar dataset, Setting 3 leads to an IMG-ROC drop across most models, approximately 10\% for PaDiM and FastFlow, and considerably larger for STFPM and RD4AD.
On the contrary, PatchCore proves to be the most resilient, showing minimal degradation and making it the most suitable model for real orbital surface analysis.
\\
Examining the results for the Mars benchmark, a different pattern emerges.
While most models such as PaDiM, RD4AD, SimpleNet, and FastFlow, sustain only marginal degradation under Setting 3, PatchCore collapses entirely, where IMG-ROC falls from 0.80 to 0.40, a trend diametrically opposed to what the lunar benchmark would suggest.
\\
We attribute this to PatchCore's memory bank mechanism: in the lunar case, anomalies are subtle and localized, introducing only mild noise when included in training. On Mars, anomalous images differ globally from normal ones, so their distinctive patches readily corrupt the memory bank, causing the observed performance to collapse.

\subsection{Qualitative Results}
\label{subsec:qualitative_results}

Another advantage of VAD models is their ability to generate anomaly maps without pixel-level annotations during training. 
Since neither the lunar nor Mars datasets provide pixel-level ground truth for the test set, we present qualitative results in Figures \ref{fig:lunar_anomaly_maps} and \ref{fig:mars_anomaly_maps}.
\\
Figure \ref{fig:lunar_anomaly_maps}.c shows anomalous lunar images overlaid with PaDiM anomaly maps. Despite being trained exclusively on normal terrain, the model accurately localizes crater structures, demonstrating open-world detection capability and confirming that VAD methods can identify unanticipated geological phenomena without supervision, a critical property for planetary exploration, where possible discoveries are unknown and cannot be anticipated at training time.
\\
Figure \ref{fig:mars_anomaly_maps}.c shows the corresponding results on Mars. Localization is comparatively less precise, which we attribute to the nature of the anomalies.
While lunar anomalies such as craters occupy compact regions against a homogeneous background, Martian anomalies, such as exposed rock interiors, mineral veins and drill holes, often span the entire image, making spatial localization inherently more challenging. Nevertheless, VAD methods achieve state-of-the-art detection performance on the Mars dataset, and the produced anomaly maps provide human operators with meaningful visual guidance to prioritize scientifically valuable observations within the tight constraints of rover tactical planning.

\begin{table}[!htb]
\centering
\caption{Computational profile of evaluated models.}
\begin{tabular}{l c c c}
\toprule
Model & \makecell{Memory \\ Footprint (MB)} & \makecell{Peak CPU \\ Memory (MB)} & \makecell{Latency \\ (ms)} \\
\midrule
PaDiM     & 10.7  & 1 & 40 \\
CFA       & 5.7  & 1 & 18 \\
RD4AD     & 116  & 3 & 41 \\
SimpleNet & 11.5  & 3 & 30 \\
PatchCore & 17.4 & 1 & 28 \\
STFPM     & 10.4 & 3 & 24 \\
FastFlow  & 8.8  & 1 & 27 \\
\bottomrule
\end{tabular}
\label{tab:models_profile}
\end{table}

\section{Conclusion}
\label{sec:conclusion}

In this work, we presented the first evaluation of VAD methods on real planetary imagery.
We introduced two dedicated benchmarks and evaluated seven feature-based VAD methods under three progressively realistic experimental settings.
Specifically, we covered orbital lunar data from the Lunar Reconnaissance Orbiter Camera and Mars surface imagery from Curiosity's Mastcam. 
\\
Our results show that feature-based VAD methods can match or exceed existing supervised and reconstruction-based baselines. On the lunar dataset, PaDiM achieves accuracy within 10\% of a fully supervised VGG-11 classifier with no label supervision.
On Mars, PatchCore surpasses the best reconstruction-based model by a substantial margin using only standard RGB channels. 
\\
Beyond detection performance, VAD methods produce interpretable pixel-level anomaly maps, a key advantage when bandwidth and operator time are scarce. 
In addition, these results are achieved with a minimal memory footprint (e.g., 10.7 MB for PaDiM), making onboard deployment on resource-constrained platforms practically viable.
\\
Taken together, our results establish annotation-free VAD as a viable approach for automated planetary analysis, combining competitive detection performance, computational efficiency, and interpretability to support tasks ranging from hazard detection and landing site selection to bandwidth-aware data prioritization.
\\
\textbf{Future Work}
This work opens several promising directions for future research.
While the Mars dataset provides six narrow-band spectral channels, our approach is limited to standard RGB by the use of pretrained backbones.
However, the additional channels carry spectral signatures invisible to RGB-only models, offering discriminative cues for identifying specific novel classes.
Therefore, future work will focus on fusion strategies to exploit this additional information.
\\
Beyond planetary surface imagery, VAD holds considerable untapped potential for a broader class of space mission-critical inspection tasks. 
Its utility could extend to monitoring solar panel integrity, detecting degradation, contamination, or impact damage before it compromises power generation. 
Similarly, rocket engine nozzle inspection could benefit from early identification of erosion patterns or structural cracks. 
Monitoring the propellant tank could flag corrosion, weld defects, or structural deformation that, if left undetected, could cause catastrophic mission failure.
Eventually, inspecting composite payload fairings could reveal delamination or micro-cracking, compromising the structural integrity.

\bibliographystyle{IEEEtran}  
\bibliography{main}

\end{document}